\documentclass{article}

\usepackage{arxiv}

\usepackage[utf8]{inputenc} 
\usepackage[T1]{fontenc}    
\usepackage{hyperref}       
\usepackage{booktabs}       
\usepackage{amsfonts}       
\usepackage{nicefrac}       
\usepackage{microtype}      
\usepackage{lipsum}		
\usepackage{graphicx}
\usepackage{natbib}
\usepackage{doi}
\usepackage{times}  
\usepackage{helvet}  
\usepackage{courier}  
\usepackage{natbib}  
\usepackage{caption} 
\usepackage{booktabs}     
\usepackage{multirow}     
\usepackage{threeparttable} 

\usepackage{amsmath,amsfonts,bm}
\usepackage{tcolorbox}
\usepackage{algorithm2e}

\usepackage{array}
\usepackage{textcomp}
\usepackage{url}
\usepackage{verbatim}
\usepackage{graphicx}

\usepackage{amssymb} 
\usepackage{multirow}
\usepackage{comment}
\usepackage[marginal]{footmisc}
\usepackage{amsfonts}
\usepackage{algorithm2e}
\usepackage{algorithmic}
\usepackage{mathtools}
\usepackage{booktabs}
\usepackage{color}
\usepackage{floatrow}
\usepackage{marvosym}
\usepackage{ntheorem}
\usepackage{url}
\usepackage{verbatim}
\usepackage{todonotes}
\usepackage{cleveref}
\usepackage{marginnote}
\usepackage{threeparttable}
\usepackage{svg}
\usepackage{subcaption}

\usepackage{floatrow}

\newfloatcommand{capbtabbox}{table}[][\FBwidth]
\usepackage{blindtext}

\usepackage{xcolor}

\SetKwComment{Comment}{/* }{ */}
\SetKwInput{KwData}{Input}
\SetKwInput{KwResult}{Output}

\newlength{\boxunit}
\newtcolorbox{lightgrey}{
  colback=gray!10, 
  colframe=gray!60,
  boxrule=0.2pt,      
  auto outer arc,     
  width=\linewidth,
  left=2.5mm,
  right=2.5mm,
  top=\boxunit,
  bottom=\boxunit,
  before skip=\boxunit,
  after skip=\boxunit
}

\newtcolorbox{greybox}{
  boxsep=0mm, 
  top=1.5\boxunit,
  bottom=1.5\boxunit,
  before skip=\boxunit,
  after skip=\boxunit,
  width=\linewidth,
  left=2.5mm,
  right=2.5mm,
}

\def\eqref#1{equation~\ref{#1}}

\def\1{\bm{1}}

\def\vc{{\bm{c}}}

\def\vh{{\bm{h}}}

\def\vv{{\bm{v}}}

\def\mA{{\bm{A}}}

\def\mD{{\bm{D}}}

\def\mG{{\bm{G}}}

\def\mK{{\bm{K}}}
\def\mL{{\bm{L}}}
\def\mM{{\bm{M}}}

\DeclareMathAlphabet{\mathsfit}{\encodingdefault}{\sfdefault}{m}{sl}
\SetMathAlphabet{\mathsfit}{bold}{\encodingdefault}{\sfdefault}{bx}{n}

\def\gE{{\mathcal{E}}}

\def\gG{{\mathcal{G}}}

\def\gL{{\mathcal{L}}}

\def\gV{{\mathcal{V}}}

\def\sR{{\mathbb{R}}}

\title{A Stage-Aware Mixture of Experts Framework for Neurodegenerative Disease Progression Modelling}

\author{
    Tiantian He,
    Keyue Jiang,
    An Zhao,
    Anna Schroder, 
    Elinor Thompson,
    Sonja Soskic,\\
    Frederik Barkhof,
    Daniel C. Alexander\\ \\
    Department of Computer Science\\
    University College London\\
    \texttt{tiantian.he.20@ucl.ac.uk}
    \\
    \texttt{keyue.jiang.18@ucl.ac.uk}
}

\renewcommand{\shorttitle}{\textit{arXiv} Template}

\begin{document}
\maketitle

\begin{abstract}

The long-term progression of neurodegenerative diseases is commonly conceptualized as a spatiotemporal diffusion process that consists of a graph diffusion process across the structural brain connectome and a localized reaction process within brain regions. However, modeling this progression remains challenging due to 1) the scarcity of longitudinal data obtained through irregular and infrequent subject visits and 2) the complex interplay of pathological mechanisms across brain regions and disease stages, where traditional models assume fixed mechanisms throughout disease progression. To address these limitations, we propose a novel stage-aware Mixture of Experts (MoE) framework that explicitly models how different contributing mechanisms dominate at different disease stages through time-dependent expert weighting. This architecture is a key innovation designed to maximize the utility of small datasets and provide interpretable insights into disease etiology. Data-wise, we utilize an iterative dual optimization method to properly estimate the temporal position of individual observations, constructing a cohort-level progression trajectory from irregular snapshots. Model-wise, we enhance the spatial component with an inhomogeneous graph neural diffusion model (IGND) that allows diffusivity to vary based on node states and time, providing more flexible representations of brain networks. We also introduce a localized neural reaction module to capture complex dynamics beyond standard processes.The resulting IGND-MoE model dynamically integrates these components across temporal states, offering a principled way to understand how stage-specific pathological mechanisms contribute to progression. When used to model tau pathology propagation in human brains, IGND-MoE outperforms purely pathophysiological and purely neural baselines in long-term prediction accuracy. Moreover, its stage-wise weights yield novel clinical insights that align with literature, suggesting that graph-related processes are more influential at early stages, while other unknown physical processes become dominant later on. Our findings highlight the necessity of designing hybrid and expert-constrained models that account for the evolving nature of neurodegenerative processes.\footnotetext{The first two authors contributed equally to this work}

\end{abstract}

\section{Introduction}

Neurodegenerative diseases exhibit a progressive propagation of pathology through the brain \cite{Busche2020SynergyDisease}. Understanding the long-term progression of these diseases from their early to advanced stages is a key challenge for developing disease-modifying treatments. However, constraints of real-world patient data acquisition often hamper such efforts. Since medical scans can be expensive or pose potential health risks, data is often collected irregularly and over a narrow time frame. Accordingly, a set of modern computational approaches, known collectively as data-driven disease progression models \cite{fonteijn2012event,young2014data}, has emerged to address the challenge of estimating population-level trajectories of change from such sparse and irregularly sampled patient data sets.

Pathophysiological disease progression models \cite{zhou2012predicting,raj2012network,seguin2023brain,garbarino2019differences, young2014data} simulate the spread of pathology over the brain using hypothetical mechanisms. These models capture spatiotemporal dynamics through two components: i) a graph that approximates the ability of each region's pathology occupancy to cause pathology appearance in each other region and ii) a mechanism of propagation between regions given that set of graph links. Network diffusion models (NDMs) \cite{raj2012network,weickenmeier2018multiphysics}, a key class of these models, assume pathology spreads by diffusing along structural brain connections from MRI. Current approaches use brain connectivity measures as proxies for graph link strength and offer interpretability. However, they face three limitations\label{limitation}: 1) \textbf{Homogeneous graph diffusion} assumes uniform diffusion rates and fixed graphs, oversimplifying brain networks and failing to model complex neurodegeneration that evolves over time~\cite{he2023coupled,Zhou2012PredictingConnectome,avena-koenigsberger2018a}; 2) \textbf{Constrained localized propagation} inadequately captures alternative processes beyond reaction-type mechanisms, such as clearance and  interactions~\cite{garbarino2021a,abi2022simulad}; and 3) The assumption of \textbf{fixed mechanisms throughout disease}, whereas contributing mechanisms likely shift during disease progression\cite{meisl2021vivo}. 

Deep learning-based time series models offer opportunities for enhancing disease progression models, leveraging their inherent flexibility and data-driven characteristics. However, existing approaches have distinct limitations when applied to neurodegenerative disease progression. Typical models for time-series analysis, such as: Recurrent Neural Networks (RNNs)\cite{rumelhart1985learning}, Long Short-Term Memory Networks (LSTMs)\cite{graves2012long}, and Gated Recurrent Units (GRUs)\cite{chung2014empirical}, discretize time series and thus exhibit limitations in handling continuous dynamical systems, irregularly sampled data, and long-term dependencies. Neural ordinary differential equations (ODES)\cite{Chen2018NeuralEquations} parameterise the derivative of the trajectory using black-box neural networks, enabling them to approximate complex dynamics with great flexibility. However, they lack interpretability, critical for clinical applications, as they cannot decompose learned patterns into mechanistic components that align with biological knowledge. Graph Neural ODEs\cite{poli2019graph} couple Neural ODEs with graph networks to model the complex dynamics evolving on the graph. However, they are rigidly bound to a predefined input graph structure that may not capture all relevant biological connections. Moreover, they cannot easily distinguish between graph-related and non-graph-related pathological processes. Thus, this type of model alone offers limited insights into how different mechanisms contribute at different disease stages or brain regions. The proper combination of those different deep learning based methods with the pathophysiological model will allow the benefit of maximising each of their expertise, and more importantly, allow us to better understand the complex mechanism of the disease progression pattern by understanding their extent of contribution at different disease stages and regions.

To address the challenge of understanding how pathological mechanisms dynamically shift over the disease course, we develop a novel stage-aware Mixture of Experts (MoE) framework that combines an existing pathophysiological model with an inhomogeneous graph neural diffusion model (IGND), where the graph diffusivity varies based on node states and time rather than being constant. Unlike traditional approaches that assume fixed mechanisms throughout disease progression, our IGND-MoE explicitly models how different contributing mechanisms dominate at different disease stages through time-dependent expert weighting. For the spatial component, we introduce an IGND parameterized by a graph auto-encoder (GAE) to address the limitation of homogeneous graph diffusion models \cite{raj2012network,weickenmeier2018multiphysics} that assume uniform diffusion rates with fixed connectivity graphs, our inhomogeneous approach allows diffusivity to vary based on both node states and time, enabling more complex dynamics where pathology spread rates adapt to disease progression. This provides a more flexible representation of network dynamics than prior approaches such as standard Neural ODEs \cite{Chen2018NeuralEquations} and Graph Neural ODEs \cite{poli2019graph}, which either lack graph structure or maintain fixed graph properties throughout the disease course. For the local component at each node, we enhance the traditional logistic growth with neural methods to increase expressiveness. By incorporating temporal attention that modulates each expert's contribution across disease stages, our model reconstructs cohort-level, long-term spatiotemporal dynamics while revealing stage-specific pathological mechanisms. This comprehensive modelling strategy not only improves predictive accuracy but, more importantly, provides new perspectives on the evolving nature of neurodegenerative processes by distinguishing whether unexplained pathological changes are inherently graph-related, non-graph-related, or emerge from their complex interactions. 

Our contributions are summarized as follows:
(1) We propose a stage-aware Mixture of Experts (IGND-MoE) framework to construct a long-term continuous disease progression trajectory from irregular snapshots. It combines pathophysiological models with neural approaches to model how different pathological mechanisms dominate at different disease stages. 
(2) We introduce an inhomogeneous graph neural diffusion model for spatial propagation and enhance localized dynamics with neural networks, both integrated through temporal attention that modulates expert contributions across disease stages.
(3) Through extensive experiments, we demonstrate our model achieves superior prediction accuracy compared to pure pathophysiological or neural approaches, particularly for long-term prediction, while providing interpretable insights into stage-specific disease mechanisms.

\section{Background: Network Diffusion Models}
\label{sec:mec_model}
The propagation of deleterious proteins in neurodegenerative disease is commonly formalized as a graph differential equation system based on the Fisher-Kolmogorov equation~\cite{Fisher-Kolmogorov-equation,Meisl2021InDisease,Raj2012ADementia,Weickenmeier2018MultiphysicsAtrophy}, where the graph is constructed based on brain connectivity. Existing literature decomposes the propagation into two principal mechanisms: (i) the spatial diffusion of toxic proteins through the brain's structural network~\cite{Raj2012ADementia} named network diffusion model (NDM); and (ii) the localized reaction (production and accumulation) of these proteins~\cite{Meisl2021InDisease}. The spatial diffusion of the protein is quantified as a \textit{homogeneous} heat diffusion process~\cite{carslaw1959conduction,DBLP:journals/tsipn/ThanouDKF17} modulated by the graph Laplacian matrix $\mL=\mD - \mA$ and rate $k$, where $\mD= \operatorname{diag}(\mA )$ is the degree matrix. The local production and aggregation process first exhibits a progressively increasing trend, ultimately stabilizing at a plateau value $\vv$, with a uniformly increasing rate $\alpha$ applicable to all regions. With the above two-component evolution, the overall process is governed by the following ordinary differential equation:
\begin{equation}
\label{eq:baseline NDM}
\frac{d \vc(t)}{d t}=\underbrace{-k [\mL \cdot \vc(t)]}_{\text{Spatial Diffusion (i)}}+\underbrace{\alpha \vc(t) \odot[\vv-\vc(t)]}_{\text{Localized Reaction (ii)}} \coloneqq f_{\text{M}},
\end{equation}
where $\odot$ is the element-wise product. Once the derivative is properly estimated, the long-term prediction is obtained by integrating the derivatives $\vc(t) = \vc(0) + \int_0^t (d\vc(t)/dt) dt$. Compared to the discrete prediction models, the PDE-based models naturally factor in the continuous dynamics along the timeline, and are thus more suitable for long-term pathology progression modeling. 


\textbf{The limitation of pathophysiological Models.} We recognise three main drawbacks:

1. \textit{The homogeneity of graph diffusion.} the model in~\cref{eq:baseline NDM} assumes that the spatial diffusion is \textit{homogeneous} with uniform node and edge characteristics, which suggests that the efficiency at which toxic proteins travel from one region to another is considered constant throughout the entire network and timeline. This hypothesis oversimplifies the actual brain network and fails to model complex neurodegeneration in real-world scenarios \cite{he2023coupled,Zhou2012PredictingConnectome,avena-koenigsberger2018a}. 
    
    
2. \textit{The constrained localized propagation.} The constrained physical models inadequately capture localized propagation, typically limited to reaction-type processes. They miss other reported mechanisms like protein clearance and interactions with other biomarkers \cite{garbarino2021a,abi2022simulad}, necessitating a more expressive modeling approach for accurate localized propagation.

3. \textit{The temporal dynamics of expert models.}  The model in ~\cref{eq:baseline NDM} only accounts for static contributions of each mechanism, however the contribution of each process to the overall toxic protein propagation varies over time. For instance, previous literature \cite{meisl2021vivo} suggests the local production takes the lead at later stages.


\section{Methodology}

\subsection{Problem Formulation}

Raw medical data are commonly obtained at the individual subject level. For each subject $i$, a longitudinal sequence of samples is provided as $\{\vc(t^{i}_s)\}_{s=1}^{S_{i}}$, where the samples are ordered chronologically for $S_i \geq 1$ distinct time points $t_s, s \in \{1, \ldots, S_i\}$. Each sample $\vc(t^{i}_s)$ constitutes an $n$-dimensional vector representing node states derived from $n$ brain regions. These brain regions define a graph $\gG = (\gV, \gE)$ on node set $\gV$ representing the brain regions, and edge set $\gE$ the connectivity between the regions, where $|\gV|=n$ and $|\gE| = e$. The adjacency matrix of the graph is represented by $\mA \in \mathbb{R}^{n \times n}$. These individual observations are then aligned to a common temporal axis representing the full spectrum of disease development. This alignment facilitates the construction of a cohort-level disease progression trajectory, specified by a sequence of node state vectors $\{\vc(t)\}_{t=1}^T$ over $T$ time points. Here, $\vc(t) = [c(u, t)]_{u \in \mathcal{V}}^\top \in \mathbb{R}^n$ is the vector where $c(u, t)$ denotes the estimated pathological quantity (e.g., protein concentration) in node $u$ at time $t$ on the global timeline. The core problem involves accurately projecting incoming individual samples onto this estimated disease timeline and predicting the subsequent evolution of the node states.

\subsection{Overview of the modelling framework.}

\begin{figure}
\centering
\includegraphics[width=\textwidth]{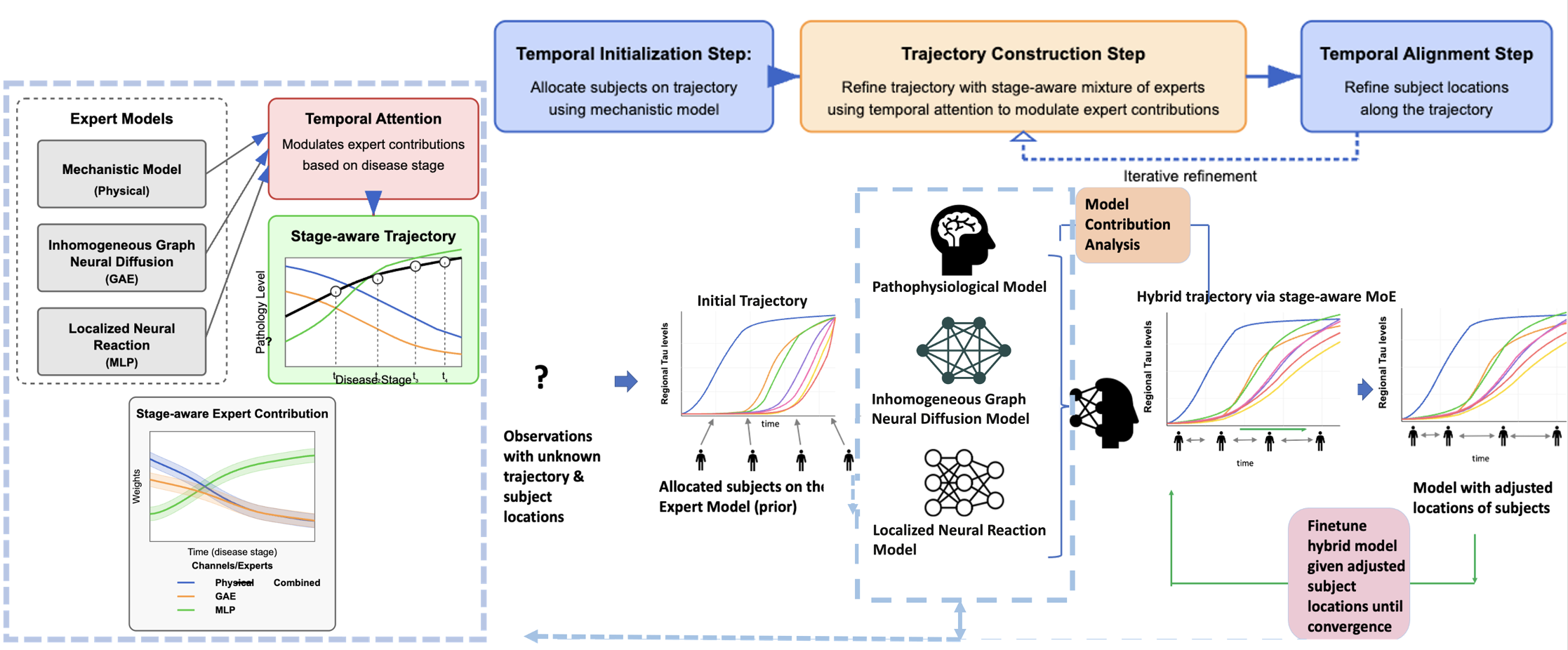}
\caption[IGND-MoE Framework Overview]{\textbf{Stage-aware Neurodegenerative Disease Progression Modeling with IGND-MoE.} This figure demonstrates the proposed IGND-MoE framework for constructing a full disease progression
process from snapshots, by iteratively carrying out the temporal alignment step for mapping each subject to the proper location on the time axis and the trajectory construction step of shaping a better trajectory through the proposed temporal-aware mixture of expert structure.}
\label{framework}
\end{figure}

\begin{figure}
\centering
\includegraphics[width=0.5 \textwidth]{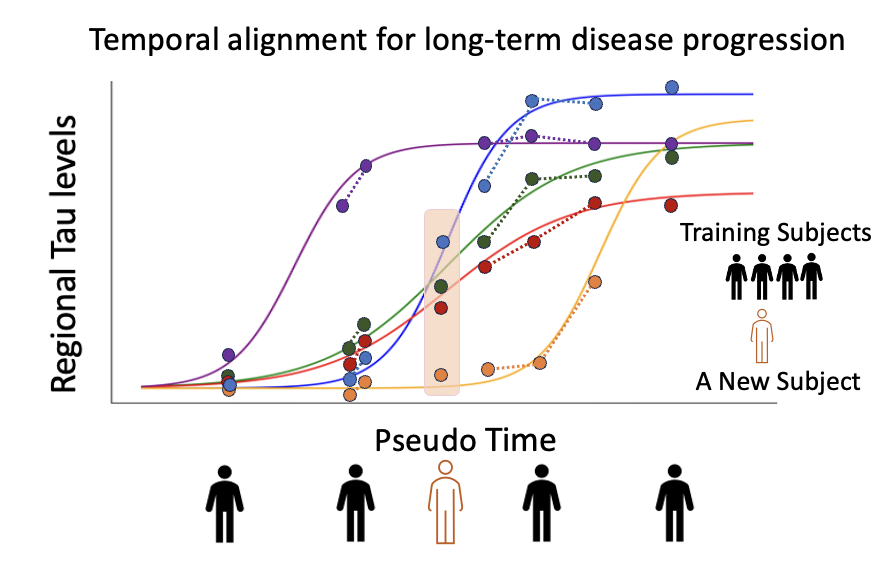}
\caption[Temporal alignment for long-term progression]{\textbf{Temporal alignment for long-term progression.} This figure visualizes how the proposed framework uses the snapshots of the individual cross-sectional data or the short longitudinal
data to construct the full disease progression trajectory. Each colour represents one brain region. The dots represent real observations. The dots connected with dashed lines represent the longitudinal observations from the same subject, where the real-time gap between the scans is available in the dataset and thus remains. The curves represent the model fitting.} 
\label{subject_align}
\end{figure}

In general, our method aims to tackle two challenges: 1) The construction of a comprehensive cohort-level disease progression trajectory from snapshots of individual-level observations (\cref{subsec:traj_const}), and 2) The development of a mixture of experts system to model how different pathological mechanisms dominate at different disease stages (\cref{subsec:model_design}). Figure \ref{framework} displays an overview of our stage-aware Mixture of Experts framework (IGND-MoE).

At the start of model training, neither the trajectory shape nor the relative location of each individual on the trajectory is known. We therefore apply a dual optimization strategy that iteratively refines both the disease trajectory and individual subject placements along this trajectory. This optimization process consists of the following steps:
\begin{enumerate}
\item \textbf{Temporal Initalization Step}: The process begins with prior knowledge about the trajectory, simulated by the pathophysiological model described by equation \ref{eq:baseline NDM}. Given this prior trajectory, each subject $i$ is allocated to the most appropriate location on the temporal axis through optimization of their pseudo time $t^i$.
\item \textbf{Trajectory Construction Step}: Given the estimated subject locations, the trajectory is refined by combining contributions from the three expert components: the pathophysiological model, the inhomogeneous graph neural diffusion, and the localized neural reaction. The temporal attention mechanism modulates each expert's contribution based on disease stage, while regularization ensures interpretability and complementarity between experts. This produces a more flexible trajectory that captures stage-specific pathological mechanisms.

\item \textbf{Temporal Alignment Step}: Given the updated stage-aware trajectory, the subject locations along the temporal axis are further refined.

\item \textbf{Iterative Refinement}: Steps 2 and 3 are repeated, starting from the previously trained model, until convergence.
\end{enumerate}

\subsection{Temporal Alignment Step} 
As displayed by Figure \ref{subject_align}, disease progression models aim to construct long-term cohort level trajectories from snapshots of individual data by estimating the location $t^{i}_{0}$ of the baseline measurement of subject $i$ on the mutual temporal axis\cite{Young2024Data-drivenBox,lorenzi2019probabilistic} so that the measurements can align as closely as possible to the assumed trajectory, by minimizing their sum squared errors:
\begin{equation}
\mathcal{L}_{f_M}\left( t^{i}_0 \right) = \sum_{i=1}^N \sum_{s=1}^{S_i} \| {c_{obs}}{(t^{i}_{0}+t^{i}_{s})}- {c}{(t^{i}_{0}+t^{i}_{s})} \|^2
\end{equation}
where $f_M$ represents the trajectory from pathophysiological model, and $t^{i}_{s}$ represents the time gap from the baseline to the $s th$ scan for subject $i$, which is recorded in the database. 

\subsection{Cohort-level Trajectory Construction Step}
\label{subsec:traj_const}

\subsubsection{Design of the stage-aware Mixture of Expert Model}

\label{subsec:model_design}

Our model combines three key components as in~\cref{eq:baseline NDM}: 1) A pathophysiological model that provides interpretable baseline dynamics based on clinical hypotheses; 2) An inhomogeneous graph neural diffusion (IGND) model that enhances spatial propagation by learning flexible, data-driven representations of brain networks; and 3) localized neural component that captures complex local dynamics beyond standard reaction mechanisms. These components are integrated through temporal attention that dynamically modulates their contributions over time. 
\begin{equation}
\label{eq:NODE}
\begin{aligned}
\frac{d \vc(t)}{d t}=\underbrace{\beta_1(t) f_{\text{M}}}_{\text{Mechanistic Model}} &+ \underbrace{\beta_2(t)f(\mA, \vc(t),t, \theta_{\text{S}})}_{f_{\text{S}}: \text{Graph Neural Diffusion}} &+ \underbrace{\beta_3(t) f(\vc(t),t, \theta_{\text{L}})}_{f_{\text{L}}: \text{Localized Neural Reaction}},\\
\end{aligned}
\end{equation}
where $\theta_\text{S}$ and $\theta_\text{L}$ are parameters for spatial diffusion and localized reaction. The mixture weights $\{\beta_j(t)\}_{j\in\{1,2,3\}}$, which satisfy $\sum_j\beta_j(t) = 1,\forall t$, dynamically modulate the contribution of each expert model throughout the disease stages.


\paragraph{Spatial Modelling with Inhomogeneous Graph Neural Diffusion.}
Traditional network diffusion models operate on homogeneous graphs, where diffusivity remains constant across all edges and throughout the entire disease progression process, corresponding to a graph diffusion process with uniform diffusion coefficients that do not change with time or node states.

\label{sec: GNDM} We extend the homogeneous spatial diffusion induced by brain connectivity to an inhomogeneous graph neural diffusion model that accounts for changes in diffusivity over states and time. In contrast to the pathophysiological model in~\cref{sec:mec_model}, it accounts for the impact of time and node states on spatial diffusion, thus enabling the modeling of more complex dynamic systems. To do so, we consider the PDE:
\begin{equation}
\frac{\partial c(u, t)}{\partial t}=\operatorname{div}[g\left(\vc(t), t\right) \nabla c(u, t)]
\end{equation}
where $\text{div}[\cdot]$ is the divergence operator and $\nabla$ is the gradient. $g(\vc(t), t)$ is the diffusivity that governs the rate at which proteins spread between the nodes. In inhomogeneous graph diffusion, the diffusivity defined over each edge varies over time $t$ and state $\vc(t)$. Stacking $g\left(\vc(t), t\right)$ will arrive in a matrix $\mG(\vc(t), t) \in \sR^{e\times e}$. Note that the process described in ~\cref{eq:baseline NDM} corresponds to the homogeneous scenario when $g(\vc(t), t) = \text{Const}$. 

To derive a discrete version of this PDE, we introduce the incidence matrix $\mK \in \sR^{n\times e}$, which indicates the connectivity between vertices and edges in the graph~\cite{DBLP:books/daglib/0037866}.
Substituting the expressions for $\text{div}$ and $\nabla$ leads to,
\begin{equation}
\begin{aligned}
\frac{\partial \vc(t)}{\partial t}&=\operatorname{div}[\mG\left(\vc(t), t\right) \mK^\top \vc(t)]\\
&=\mK\cdot\mG\left(\vc(t), t\right)\cdot \mK^\top \vc(t)
\end{aligned}
\end{equation}
However, modeling an inhomogeneous diffusion that refines the graph requires considering all possible edges, resulting in a quadratic computational complexity with $e = n^2$. Since $\mG$ is symmetric, $\mK\cdot\mG\left(\vc(t), t\right)\cdot \mK^\top$ is positive semi-definite and can be approximated by a low-rank factorization. This motivates us to learn two matrices $\mM_\text{Enc}, \mM_\text{Dec}\in \sR^{n\times e^\prime}, e^\prime \ll e$ depending on states and time such that,
\begin{equation}
    \mM_\text{Dec} \cdot \mM_\text{Enc}^\top \approx \mK\cdot\mG\left(\vc(t), t\right)\cdot \mK^\top.
\end{equation} 
The problem is equivalent to finding the low-rank latent representation $\vh(t)$ that correctly predicts $d \vc(t)/dt$, which inspires us to use an auto-encoder-like model. To this end, GAE naturally excels in modeling inhomogeneous graph diffusion given its ability to generate latent representation and refine graph structure. The encoder maps the input graph $\gG$ with node features $\vc(t)$ into a latent representation $\vh(t)$ via graph convolutional layers. Specifically, for node $u$, the encoder calculates the latent representation as,
\begin{equation}
\vh(u,t)=\text{Enc}(c(u,t), \bigoplus_{v \in [N]} \text{Prop}\left(c(u, t), c(v,t), \mA_{uv}\right)),
\end{equation}
where $\text{Enc}(\cdot)$ and $\text{Prop}(\cdot)$ are encoder and propagation functions respectively, $\oplus$ is a permutation-invariant aggregation function and $[N] = \{1, \ldots, N\}$. Moreover, the decoder refines the graph structure and predicts the node signal by,
\begin{equation}
\frac{d\vc(t)}{dt} = \text{Dec}(\vh (t), \hat{A}), \quad \text{with }  \hat{A} =\sigma\left(\vh(t) \vh(t)^{\top}\right).
\end{equation}
$\text{Dec}(\cdot, \cdot)$ is the decoder function that takes in the latent representation of refined graph structure, and $\sigma(\cdot)$ is the normalization function to make sure the refined adjacency matrix is valid. With such a design, the graph auto-encoder provides an estimation of the derivative of the pathology propagation trajectory while refining the graph structure.

\paragraph{Localized Neural Reaction.}
\label{sec: NODE}In addition to spatial diffusion of pathological proteins via brain connections, their localised replication plays an important role in pathology accumulation. To enhance the expressiveness of this process, we use a multi-layer perception (MLP) to approximate such a process:

\begin{equation}
f_\text{L} = \frac{d \vc(t)}{d t}= \text{MLP}(\vc(t), \theta_{\text{L}}).
\end{equation}

\subsubsection{Training of the Trajectory Construction Step}
Beyond its enhanced modeling capacity, the proposed framework provides an interpretable way to assess the contribution of each component and identify pathological patterns that the pathophysiological model may have overlooked. We achieve this by incorporating norm and orthogonal regularizations to diversify the model outputs and disentangle the impact of each individual component.
The overall IGND-MoE model is trained by the following loss:
\begin{equation}
    \gL = \gL_{\text{traj}} + \lambda_1\gL_{\text{norm}} + \lambda_2 \gL_{\text{ortho}}
\end{equation}
where $\gL_{\text{traj}}$ is the ODE trajectory loss, $\gL_{\text{norm}}$ is the norm loss on the learning-based model output and $\gL_{ortho}$ is the orthogonal loss on each model's output. $\lambda_1$ and $\lambda_2$ are the hyper-parameters to balance three loss terms. Specifically, $\gL_{\text{traj}}\left(\theta_\text{S}, \theta_\text{L}, \theta_\text{M}\right)=\sum_{i=1}^N \sum_{s=1}^{S_i} \| c_\text{obs}{(t^{i}_{s})}- {c}{(t^{i}_{s})} \|^2$ with $c_\text{obs}{(t^{i}_{s})}$ is the observation at $t^{i}_{s}$, where $i$ ranges from $i$ to $N$  and $s$ ranges from 1 to $S_i$. The norm loss is used to prioritize the contribution from the pathophysiological model and use the learning-based model as a complement to encourage interpretability and improve generalization ability. Similar to~\cite{yin2021augmenting}, we regularize the training through the Frobenius norm of model outputs $\gL_{\text{norm}} = \|f_\text{S}\|_\text{F} + \|f_\text{L}\|_\text{F}$. 
  
To encourage the diversity among the models’ predictions, we add a orthogonal regularization term that penalizes the correlation between the outputs of individual ensemble members. This forces each model to learn complementary features or focus on different aspects of the data, which can lead to better generalization and improved overall performance once combined. $$\gL_{\text{ortho}}=\sum_t\lambda \sum_{p\neq q}\left(\tilde{f}_p(t) \cdot \tilde{f}_q(t)\right)^2,  \tilde{f}_p=f_p-\text{mean}(f_p),$$
with $p$ and $q$ index the models and the mean is taken across the subjects.

\section{Experiments and Results}

In this section, we will evaluate the proposed model in the task of long-term disease progression of tau pathology in human brains. First, we will benchmark our model against established spatio-temporal approaches, including discrete models and continuous models with both homogeneous and inhomogeneous configurations. Subsequently, we will conduct in-depth clinical analyses by examining how different component contributions evolve throughout disease progression. We will also identify the specific brain regions and disease stages where our model demonstrates improvement by comparing error maps between our proposed model and classical pathophysiological models. The results will provide insights into the synergistic effects of various model components (each representing distinct physical processes) within the brain's dynamic system. We also demonstrate that with proper constraints from the expert, the proposed method is more robust in the out-of-distribution case (OOD). Please refer to Appendix A.1.

\subsection{Experimental Setups}
\textbf{Data.} 
In this study we model the dynamics of tau protein, which we compare to standardized uptake value ratios (SUVRs) obtained from PET imaging from the Alzheimer’s Disease Neuroimaging Initiative (ADNI) database (adni.loni.usc.edu) \cite{landau_flortaucipir_2021}. The SUVRs of cortical regions were normalised to [0,1] across all participants and regions. For each cortical region, we implement a two-component Gaussian mixture model to establish a cutoff for tau-positivity as the mean of the negative distribution plus its one standard deviation. The cohort consists of N = 216 individuals (378 observations) who had positive amyloid-beta and tau status. \textbf{Apart from ADNI, we also carry out validation on two external datasets} - the Harvard Ageing Brain Study (HABS) and the Anti-Amyloid Treatment in Asymptomatic Alzheimer's Disease (A4) study. Please see Appendix A.2 for detailed results. Appendix A.3 displays more detailed data information.

The structural connectome is the group average across 50 individuals from the Microstructure-Informed Connectomics Database \cite{Royer2022AnNeuroscience} defined by the Desikan-Killiany Atlas \cite{desikan_automated_2006}, to formulate a stable and continuous cohort-level trajectory that captures the average disease progression.

\textbf{Training methodology}
We implement cross-validation by randomly assigning 35 subjects each to validation and test sets, with the remaining subjects forming the training set. All longitudinal scans from the same subject are kept together in their assigned sets, preserving the actual time intervals between measurements. The validation step happens after an epoch of trajectory optimization on the training data, i.e. the subjects from the validation set are allocated on the trajectory from each training epoch through stage optimization. Finally, the relative location of the subjects from the test set is estimated, and the corresponding model performance is recorded as the test metric.

\textbf{Benchmarks.} \label{benchmarks} We compare our IGND-MoE with three types of models: 1) temporal-GNN models, which are usually combinations of traditional discrete time series models (RNN, LSTM and GRU) and classic GNN models for temporal modelling \cite{DBLP:conf/iclr/LiYS018,DBLP:journals/corr/abs-1905-11485,DBLP:conf/iconip/SeoDVB18}. Since these models by nature cannot handle smooth and continuous case, we apply kernel interpolation methods to make them comparible with our proposed models, see Appendix for details.  2) Neural ODE models designed from homogeneous spatial diffusion\cite{zang2020neural} 3) Hybrid model with MoE structure but we replace the backbone model with other GNN\cite{DBLP:conf/iclr/KipfW17} models, such as GAT\cite{DBLP:journals/corr/abs-1710-10903}, g-transformer\cite{DBLP:conf/eccv/YuMRZY20}, etc. And we also consider the case when we remove the stage-aware characteristics of the mixture mechanisms and thus make the model contribution time-independent.

\textbf{Metrics.} \label{metrics} We evaluate model performance using two metrics on the test set: sum of squared errors (SSE), and average Pearson correlation coefficient (R). SSE is calculated across all subjects and regions of interest to quantify prediction accuracy. For each observation, we compute the Pearson correlation coefficient between predicted and measured tau SUVR signals across brain regions, which captures the spatial distribution similarity of pathology. These individual correlations are then averaged to obtain a group-level measure of the model's ability to reproduce region-specific pathology patterns independent of absolute magnitude.

\subsection{Quantitative Model Comparison}

We conduct the experiments using models mentioned in \ref{benchmarks}. Table~\ref{tab: quant_result} compares models on three metrics described in \ref{metrics}. With our IGND-MoE, a consistent improvement is found in all metrics on test data, in comparison to the short-term discrete and homogeneous graph diffusion methods. Specifically, 1) When replacing the GAE with other GNNs, the performance drops. The results suggest that GNNs without any graph refinement techniques lack the capacity to fully capture the complex dynamics. 2) When replacing temporal attention and assuming fixed contribution from experts, the model performance also drops. \textbf{This shows that the contribution from different disease mechanisms varies at different disease stages}.


\begin{table}[htbp]
\centering
\scriptsize
\resizebox{\textwidth}{!}{
\begin{threeparttable}
\renewcommand\arraystretch{0.85}
\begin{tabular}{llccc}
\toprule 
& \multirow{2}{*}{\textbf{Method}} & \multirow{2}{*}{\textbf{Backbone Model}} 
& \multicolumn{2}{c}{\textbf{Metrics}} \\
\cmidrule(r){4-5}
&&& Test SSE & Test R Corr \\
\midrule
& Physical Model~\cite{Raj2012ADementia}
& / & 17.76 $\pm$ 0.32 & 0.657 $\pm$ 0.018 \\
\midrule
& \multirow{3}{*}{Discrete Methods with interpolation} 
& LSTM-GNN~\cite{DBLP:journals/corr/abs-1905-11485} & 17.99 $\pm$ 1.15 & 0.649 $\pm$ 0.008 \\
&& RNN-GNN~\cite{DBLP:conf/iclr/LiYS018} & 21.34 $\pm$ 1.82 & 0.610 $\pm$ 0.020 \\
&& GRU-GNN~\cite{DBLP:conf/iconip/SeoDVB18} & 15.66 $\pm$ 0.76 & 0.676 $\pm$ 0.020 \\
\midrule
& \multirow{3}{*}{\begin{tabular}[c]{@{}l@{}}Homogeneous Graph Diffusion\\~\cite{zang2020neural}\end{tabular}}  
& NeuralODE (localized) & 19.65 $\pm$ 0.27 & 0.701 $\pm$ 0.002 \\
&& NeuralODE (cross-node) & 16.73 $\pm$ 0.20 & 0.700 $\pm$ 0.002 \\
&& NDCN & 14.89 $\pm$ 0.19 & 0.677 $\pm$ 0.014 \\
\midrule
& \multirow{4}{*}{Inhomogeneous Graph Diffusion}  
& MoE-GCN & 16.50 $\pm$ 1.04 & 0.676 $\pm$ 0.019 \\
&& MoE-GTrans & 16.30 $\pm$ 0.83 & 0.679 $\pm$ 0.049 \\
&& MoE-GAE (w/o Temporal) & 15.55 $\pm$ 0.44 & 0.681 $\pm$ 0.020 \\
&& MoE-GAE (ours) & \textbf{13.97 $\pm$ 0.30} & \textbf{0.717 $\pm$ 0.014} \\
\bottomrule
\end{tabular}
\caption{Quantitative results for long-term disease trajectory prediction.}
\label{tab:quant_result}
\begin{tablenotes}
\item The results are averaged over 3 trials using three random seeds and the mean \& standard deviation are reported.
\end{tablenotes}
\end{threeparttable}
}
\end{table}

\subsection{Qualitative Results Analysis}

\begin{figure}
  \centering
  \label{combine}
    \centering
    \includegraphics[width=\textwidth]{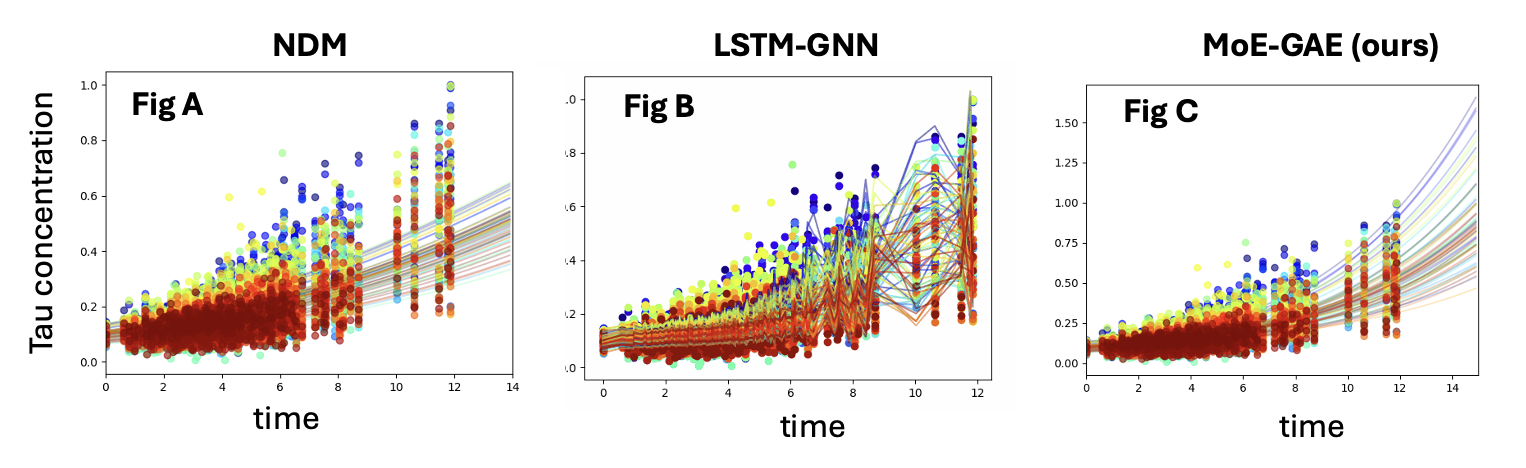}
    \caption{\textbf{Model predictions}. Each curve depicts tau‐accumulation trajectories for one of 68 cortical regions, with dots marking individual observations.}
    \label{visuals}
  \hfill
  \label{fig:combined}
\end{figure}

\begin{figure}
\centering
\includegraphics[scale = 0.15]{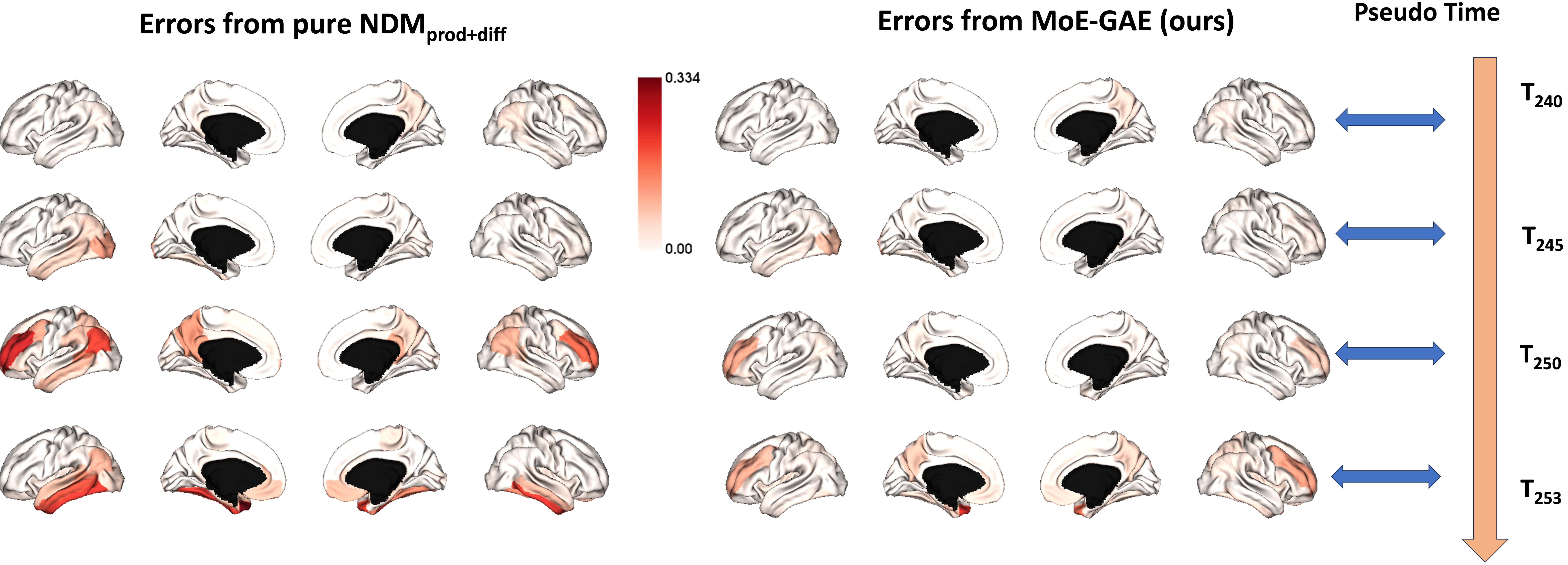}
\caption[Regional Error mapping on the brain]{\textbf{Regional Error mapping on the brain}. The plot displays the distribution of the error pattern during disease progression from the real observations using the physical model and the MoE-GAE model. The colour bar, shared by all brain plots, displays the extent of error levels.} 
\label{brain_errs}
\end{figure}

With our proposed framework, we construct the long-term trajectory of tau propagation in Alzheimer's disease using a mixture of different experts. Figure \ref{visuals} visualises the optimisation trajectories obtained from three different models on training data. 
Figure\ref{visuals}: A shows that noticeable patterns cannot be described at the later disease stages using the pure pathophysiological model. Figure \ref{visuals}: B displays that without proper constraints, data-driven deep learning models like LSTM-RNN easily overfit. Figure \ref{visuals}: C shows that the pathophysiological component can give a more plausible constraint to the MoE trajectory, and the proposed model structure, with the proper choice of the additional component, can describe more unknown patterns.

A key insight is the dynamic contribution from each expert model, dependent on disease stage.  The temporal attention, which represents the model's contribution versus disease stage, shows the physical model and GAE contributions decreasing with disease progression, while the MLP becomes dominant at later stages. This suggests graph-related processes are more influential early in our cohort, while other unknown physical processes dominate later stages. The existing contribution from GAE indicates that network diffusion isn't the only graph-related mechanism. These findings align with literature showing pathophysiological model pathology diffusion occurs primarily in earlier disease stages with deep learning models complementing later stages where other mechanisms become more prominent\cite{meisl2021vivo}. The minority pattern during some kind of data split might hint at the existence of more than one subgroup. Figure \ref{brain_errs} demonstrates the regional error of each model with disease progression, focusing on the later stages, from which we can gain insights about each region and time. The original pathophysiological model provides poor understanding and is complemented by the proposed MoE-GAE model, as shown by the higher regional errors in the left panel.

\section{Discussions and Conclusions}

We propose a novel framework for modelling long-term disease progression by combining pathophysiological models with neural networks through our IGND-MoE approach. This mixture of experts effectively characterises the evolving dynamics of tau propagation across different disease stages, providing interpretable insights into disease mechanisms. Our results reveal that graph-related processes are more influential early on, while other unknown physical processes become more prominent later, findings that align with existing literature. The IGND-MoE model achieves superior long-term prediction accuracy and provides robust out-of-sample predictions. The inherent limited number of longitudinal tau-PET data, a well-known constraint in this field \cite{leuzy2023comparison, yang2021longitudinal}, presents challenges for researchers in this field. However, our framework is specifically designed to address this by constructing cohort-level trajectories from sparse observations and mitigating the risk of overfitting through our expert-constrained architecture with special loss design. The use of a group-averaged structural connectome, a common approach for modelling cohort-level progression, is also implicitly refined by our model's Graph Auto-Encoder (GAE) component, which refines a data-driven propagation graph from individual features, adapting to individual pathology patterns. Our findings, validated across multiple independent datasets, demonstrate that this complex but carefully constrained approach provides a robust and generalizable proof-of-concept. As additional data becomes available, our method can be further refined to enhance its clinical utility, including its adaptability to other biomarkers and diseases.

\bibliographystyle{plainnat}

\bibliography{refs.bib}
\clearpage

\end{document}